\documentclass{article}
\usepackage[preprint]{tmlr}

\usepackage[utf8]{inputenc}
\usepackage[T1]{fontenc}
\usepackage{amsmath,amssymb}
\usepackage{CJKutf8}
\usepackage{graphicx}
\usepackage{booktabs}
\usepackage{hyperref}
\usepackage{url}

\title{Calibrated Surprise: An Information-Theoretic Account of Creative Quality}

\author{\name Bo Zou \email \AND \name Chao Xu \email }

\begin{document}
\maketitle

\begin{abstract}
In the era of large language models, creative writing quality lacks a computable theoretical anchor. The dominant approaches are rubric scoring --- decomposing holistic aesthetic judgment into sub-scores --- and RLHF preference signals --- replacing quality with group votes. Both bypass the statistical structure of the text itself. This paper provides an information-theoretic foundation to fill this gap.

We propose \textit{calibrated surprise} as the information-theoretic essence of excellent creative writing. This judgment matches reading intuition and covers its opposite. Each of the four combinations maps to a clear quality verdict:

\begin{center}
\small
\begin{tabular}{@{}lp{5.5cm}p{5.5cm}@{}}
\toprule
 & \textbf{Calibrated} & \textbf{Uncalibrated} \\
\midrule
\textbf{Surprising} & \textbf{Excellent} --- well-constrained, rare choice & Noise --- surprising but unsupported by constraints \\
\textbf{Unsurprising} & Formulaic --- correct but predictable & Mediocre --- neither constrained nor informative \\
\bottomrule
\end{tabular}
\end{center}

This discriminative power holds at both the macro scale (complete story, core architecture, etc.) and the micro scale (a single line of dialogue, a single punchline, etc.). This paper focuses on the information-theoretic model and does not develop the literary-theoretic argument here. \S~8.3 treats humor as a representative case.

This literary judgment admits a precise mathematical formulation. Under full-dimensional constraints $Y$, feasible writing choices are forced into an extremely narrow space. The rare survivors are, from the unconstrained perspective, exactly the least predictable choices. Both are measured precisely by Shannon mutual information $I(X;Y) = H(X) - H(X|Y)$ --- ``calibrated'' corresponds to $H(X|Y) \to 0$; ``surprising'' corresponds to $H(X)$ going high. The subtraction structure of the formula naturally separates ``well-grounded surprise'' from ``pure noise.''

\textbf{Terminology note}: In this paper, \textit{calibrated} follows its standard technical meaning in statistics and machine learning: precise alignment between a probability/constraint framework and the true distribution (as in \textit{calibrated classifier} or \textit{calibration error}), emphasizing exact satisfaction of constraints rather than subjective emotional calibration. \textit{Surprise} takes its information-theoretic meaning: a low-probability event with high $-\log P(x)$, not emotional astonishment. The specific technical content of both terms is detailed in \S~1.3 and \S~2.1.

From the same mathematical structure, mediocrity is characterized as ``the absence state of full-dimensional chain-calibration.'' Excellence and mediocrity are not two things that need separate specification. They are the dual products of the same set of constraints at different satisfaction levels.

This paper establishes two pillars. \textbf{Static pillar}: when ethos (character), mythos (plot), lexis (diction), dianoia (theme), and other full-dimensional constraints are imposed simultaneously, writing choices that satisfy all constraints collapse sharply in number. The surviving solutions are necessarily low-probability. \textbf{Dynamic pillar}: the chain rule of information theory captures how each choice is constrained by prior choices and imposes new constraints on what follows. Macro-level decisions naturally contribute a larger share of total information in the chain sum because they constrain a larger subsequent space. The long-standing question of ``how to weight macro vs. micro'' in quality assessment is dissolved as a direct mathematical corollary of the chain rule, with no ad hoc weighting required.

The direct engineering implication is: improving LLM creative quality judgment is equivalent to calibrating the internal conditional probability distribution $P(x|y)$ to systematically reduce conditional entropy $H(X|Y)$. Specific alignment protocols, training data production, and evaluation benchmarks are natural follow-on directions from this anchor. This paper does not undertake them; it provides their theoretical foundation.

We use token-level logprobs from Qwen1.5-7B as an operational proxy for the ideal reader's probability distribution. Across 20 pairs (12 Chinese / 8 English) of high-quality vs. systematically degraded literary passages, 20/20 pairs support the core prediction: high-quality passages have systematically higher $I(X;Y)$ than their degraded versions. \S~5.5 actively declares the proxy boundary between the logprob-based estimator and the theoretical $Y$.

External independent validation: Aristotle's law of probability and necessity, Gardner's fictional dream, Shklovsky's defamiliarization --- literary critical observations spanning two thousand years --- each takes its position at a different location within the same information-theoretic structure. This provides a second type of evidence entirely independent in nature from the computational evidence.
\end{abstract}

\section{Introduction}

\subsection{The Vacuum in LLM Creative Quality Measurement}

With large language models now deployed at scale, ``how to measure creative writing quality'' has become an engineering question that urgently needs an answer. Yet the two mainstream approaches both bypass the statistical structure of the text itself.

\textbf{The first approach is rubric scoring}. The AI alignment field tries to decompose creative quality into sub-dimensions scored independently --- for example, Scale AI's Rubrics as Rewards framework decomposes quality into independently scored components. This approach achieves quantifiability at the engineering level, but it systematically loses holistic aesthetic judgment --- it uses the wrong instrument. Recent empirical work has begun to confirm this problem: LLMs show systematically higher agreement with human experts on holistic evaluation than on dimension-by-dimension rubric evaluation \citep{zheng2023judging, liu2023geval}.

\textbf{The second approach is RLHF preference signals} --- replacing quality judgment with group votes. This degrades ``what makes a work good'' to ``which version more people checked.'' In a population of receivers whose aesthetic sophistication follows some distribution, this approach systematically pulls the quality target toward the population mean. Truly excellent creative work, however, tends to live in the tail of the distribution.

Both approaches share a fundamental problem: neither treats quality as \textbf{a structural property of the text's own symbol sequence}. Instead, both use indirect proxy signals --- rubric scores are an analyst's subjective decomposition; preference votes are a statistical average of the receiving end. This paper provides neither another rubric system nor another preference aggregation scheme. It provides a unified information-theoretic measure.

The motivation for this measure comes from a judgment that has logical prior necessity: \textbf{whatever a creative work ultimately delivers to its audience is a symbol sequence; if quality differences exist in the work, they must be structural differences in the symbol sequence; information theory, as the foundational mathematical theory of the statistical structure of symbol sequences, has natural jurisdiction over these differences.} This judgment requires no empirical support. It is given purely by logic: either quality differences are encoded in the statistical structure of the symbol sequence --- and then information theory can capture them --- or quality differences are not encoded in the symbol sequence --- which would mean two texts with identical content could be one excellent and one mediocre. That is absurd.

The real question has never been ``whether information theory can explain creative quality.'' The question is ``which quantity in information theory corresponds precisely.'' This paper's contribution is to answer the latter. The next section gives the answer; \S~2–\S~5 provide the theoretical argument, external validation, and computational evidence.

\begin{quote}
The core claim of this paper is grounded in the authors' long-term systematic observation of creative practice (see \S~1.3 and \S~6 for details). But its validity does not depend on the credibility of that source --- any reader can independently verify the logical chain and computational evidence in \S~2–\S~5.
\end{quote}

\subsection{This Paper's Proposal: A Computable Theoretical Anchor for LLM Creative Quality Alignment}

\begin{quote}
\textbf{Core claim}: Mutual information $I(X;Y) = H(X) - H(X|Y)$ is the computable theoretical anchor for LLM Creative Quality Alignment (CQA). It replaces the current approaches --- rubric scoring and RLHF preference signals --- with a direct calibration target on the model's internal conditional probability distribution $P(x|y)$. CQA is no longer an engineering assembly problem of ``how to quantify aesthetics.'' It is a problem with a clear mathematical definition: \textbf{maximize $I(X;Y)$ under full-dimensional constraints $Y$, which is equivalent to concentrating probability mass on calibrated writing choices}.
\end{quote}

The mutual information framework established in this paper defines creative quality as a problem observable through the model's probability distribution. Excellent creative writing is represented mathematically by high mutual information $I(X;Y)$ --- conditional entropy $H(X|Y)$ approaching zero under full-dimensional constraints $Y$, while unconditional entropy $H(X)$ goes high. The direct engineering implication of this definition is: improving LLM creative quality judgment is equivalent to calibrating the internal conditional probability distribution $P(x|y)$, so that the model concentrates probability on truly calibrated writing choices after receiving full-dimensional constraints --- rather than relying on rubric scores or RLHF preference signals as low-bandwidth indirect feedback.

This paper focuses on establishing the theoretical anchor. Alignment protocols, training data production, empirical fine-tuning, and evaluation benchmarks based on this anchor are natural follow-on directions. This paper does not undertake them.

\subsection{Core Claim and the Definition of Calibration}

This paper's core claim:

\begin{quote}
\textbf{The essence of excellent creative writing is ``calibrated surprise'' --- full-dimensional constraints force feasible solutions into an extremely narrow space, and the rare survivors are exactly the least predictable choices from the unconstrained perspective.}
\end{quote}

``Calibrated'' corresponds to $H(X|Y) \to 0$: under full-dimensional constraints, feasible solutions are forced into an extremely narrow space. ``Surprising'' corresponds to $H(X)$ going high: these rare survivors are extremely rare from the unconstrained perspective. The two are not separate goals to pursue independently. They are two sides of the same set of constraints. The mutual information formula $I(X;Y) = H(X) - H(X|Y)$ is the precise measure of ``calibrated surprise.''

This directly yields a mathematical corollary: \textbf{full-dimensional creative calibration and mediocrity are mutually exclusive.} Creative solutions that simultaneously pass all constraints are extremely rare; they are necessarily rare and non-mediocre from the unconstrained perspective. This mutual exclusion does not need human judgment to grant it legitimacy. It is a mathematical necessity of the core claim.

The term ``calibrated'' needs a precise definition. In this paper, calibration means the convergence of author intent, reader's reasonable expectation, and the logic of reality --- three fully independent sources of judgment arriving at consistent positive conclusions about the same work. ``Calibrated'' does not mean technically correct. The well-made play (la pièce bien faite) --- the highly formulaic dramatic genre pioneered by the nineteenth-century French playwright Eugène Scribe --- can be perfectly correct on the mythos (plot) dimension. But its hollowness on ethos (character) and dianoia (theme) makes it far from ``full-dimensionally calibrated.'' The sufficient condition for the mutual exclusion corollary is ``full'' --- partial calibration cannot guarantee non-mediocrity.

It is important to note that ``surprising'' in this paper strictly refers to the information-theoretic $-\log P(x)$ --- the low-probability character of a writing choice under the reader's default distribution --- not everyday ``astonishment'' or emotional surprise. Under this strict definition, ``surprising'' can take many forms: a humorous punchline, a narrative reversal, Shklovsky's defamiliarization --- all creative techniques are specific instances of the same information-theoretic structure of ``calibrated surprise,'' at different time scales and different levels of abstraction. The English term \textit{calibrated surprise} takes \textit{surprise} in this technical sense, not as an emotional experience.

The film \textit{The Godfather}'s wholesale reframing of ``criminals'' as a category is a classic ``calibrated surprise'' in cinema history. The film relocated criminals from the reader's default distribution of ``vile/contemptible'' to the low-probability region of ``dignified/looked up to.'' At the time, this creative decision was a massive surprise for audiences. Its influence persists today --- in both literature and film, portraying criminals as ``dignified/looked up to'' is no longer unusual. It has been absorbed by countless subsequent works into the new default distribution.

In this sense, the pursuit of calibrated surprise is the creative artist's ultimate mission.

\begin{quote}
\textbf{Terminology note}: The example above uses a film (\textit{The Godfather}). The core argument of this paper takes narrative literary fiction as its primary subject. However, the information-theoretic structure of ``calibrated surprise'' applies equally to literary and cinematic narrative --- both share the same underlying logic at the level of creative decisions (the intersection structure of full-dimensional constraints such as character setting, plot causality, and thematic coherence is fully isomorphic). The difference is only in the delivery medium (text vs. audio-visual). Therefore, this paper does not distinguish between literary and cinematic examples. Wherever film examples appear, the quality analysis of creative decisions is equivalent to that for literary works.
\end{quote}

\subsection{The Analytical Tool and the Receiver Premise}

This section covers two preliminary matters. First, mutual information is selected as the core analytical tool. Second, this paper anchors quality to the probability judgment of the ``ideal reader'' --- what we call the receiver premise. Together these form the minimal framework on which the arguments in \S~2–\S~5 rest.

Shannon mutual information provides the precise analytical tool for the above claim:

\[ I(X;Y) = H(X) - H(X|Y) \]

where $X$ is the writing choice and $Y$ is the intersection of full-dimensional constraints facing that choice. The probability distribution $P(x)$ is anchored to the ideal reader's judgment (see \S~6 for details). $I(X;Y)$ measures the degree to which a writing choice is ``forced'' by reality rather than randomly generated. When full-dimensional constraints force the choice toward an almost unique solution ($H(X|Y) \to 0$), that solution is necessarily rare from the unconstrained perspective ($H(X)$ rises automatically) --- ``calibrated'' automatically produces ``surprising.''

This formula also handles the dynamic structure. Creative writing is not a static object. Each choice is constrained by what came before and creates new constraints for what follows. The chain rule of information theory reveals that macro-level decisions naturally contribute a larger share of total information because they constrain a larger subsequent space. No ad hoc weighting scheme is needed.

Before entering the theory, a philosophical premise must be established. If no content consumer existed anywhere in the universe, the quality ``good'' or ``bad'' of a work would lose all grounding --- the distinction between ``excellent'' and ``mediocre'' cannot stand without a receiver. This paper therefore anchors quality measurement to the probability judgment of the ``ideal reader,'' not to isolated properties of the text. This position is in the tradition of reception aesthetics \citep{jauss1982aesthetic, iser1978act}, but this paper provides a precise formulation at the information-theoretic level.

A clarification is needed here to avoid an apparent conflict with \S~1.1's claim that ``quality naturally resides in the statistical structure of the symbol sequence.'' This paper distinguishes two layers: \textbf{measurement structure} and \textbf{realization condition}. \textbf{(i) Measurement structure} --- mutual information $I(X;Y) = H(X) - H(X|Y)$ as a mathematical object exists independently of any specific receiver. This is the objectivity discussed in \S~1.1. \textbf{(ii) Realization condition} --- for the quality value of any specific work to be actually computed, both $H(X|Y)$ and $H(X)$ need an actual carrier of the probability distribution (the ideal reader as the source of probability judgment). The former is the objectivity of the definition; the latter is the instantiation condition for measurement. This distinction is analogous to: the gravitational formula $F = GmM/r^2$ is independent of objects, yet the realization of ``force'' requires objects --- the structure has no observer dependence; the realization does. There is no contradiction.

\subsection{Contributions}

\begin{enumerate}
\item Proposing ``calibrated surprise'' as the information-theoretic description of the essence of excellent creative writing, and proving its mathematical corollary --- full-dimensional creative calibration and mediocrity are mutually exclusive. This is the theoretical foundation of this paper.
\item \textbf{Providing a computable theoretical anchor for LLM creative quality alignment}: based on the above theoretical foundation, recasting the goal of CQA from ``replicating group preference'' or ``satisfying a rubric'' to ``maximizing $I(X;Y)$ under full-dimensional constraints $Y$'' --- a target directly computable from model logprobs. This is the core contribution of this paper to the ML/alignment community.
\item Operationally defining ``creative calibration'' as the convergence of author intent, reader's reasonable expectation, and the logic of reality.
\item \textbf{Static pillar}: full-dimensional constraint accumulation leads to collapse of the solution space; $H(X|Y) \to 0$ automatically produces high $H(X)$ --- calibration mathematically entails non-mediocrity.
\item \textbf{Dynamic pillar}: under the chain rule, each writing choice is both constrained by prior choices and imposes new constraints on subsequent ones; macro-level decisions naturally contribute a larger share of information because they constrain a larger subsequent space. The macro/micro weighting problem is thus resolved naturally, with no ad hoc patch.
\item \textbf{Computational evidence}: using token-level logprobs from Qwen1.5-7B, verifying the core prediction on 20 pairs (Chinese/English) of high-quality vs. systematically degraded passages, with 20/20 consistent results. \S~5.5 actively declares the proxy boundary between the logprob-based estimator and the theoretical $Y$.
\item \textbf{External independent validation}: unifying Aristotle's law of probability and necessity, Gardner's fictional dream, Shklovsky's defamiliarization, and other literary critical observations scattered across two thousand years under different positions in the same information-theoretic structure --- providing a second type of evidence entirely independent in nature from the computational evidence.
\end{enumerate}

\section{Theoretical Foundation}

\subsection{The Mutual Information Formula}

The core analytical tool of this paper is Shannon mutual information:

\[ I(X;Y) = H(X) - H(X|Y) \]

In the context of creative quality analysis:

\begin{itemize}
\item $X$: writing choice --- the specific choice the author makes at a decision point in the narrative.
\item $Y$: the intersection of full-dimensional constraints (see \S~2.3).
\item $H(X)$: information entropy of the choice --- how surprising this choice is from the perspective of an external observer without background knowledge.
\item $H(X|Y)$: conditional entropy --- the remaining uncertainty in the choice after full-dimensional constraints are fully given.
\item $I(X;Y)$: mutual information --- the actual information transmitted between the writing choice and the constraints, i.e., surprise minus noise.
\end{itemize}

$I(X;Y)$ measures the degree to which this writing choice is ``forced'' by the constraints rather than generated randomly.

The film \textit{L.A. Confidential} (1997) provides a classic example. Two detectives, Jack and Ed, visit a bar to follow up on a lead. While Ed is asking questions, he makes a sarcastic remark to a woman present --- he has concluded she is a call girl who had plastic surgery to look like a movie star. (The film has earlier established that there is a prostitution ring in Los Angeles that uses ``looking like a celebrity'' as its selling point.) The audience at this moment is also led astray by Ed's judgment.

Mapping this passage to the formula:

\begin{itemize}
\item $X$ (writing choice): Ed mistakes the actual movie star for a surgery call girl.
\item $Y$ (full-dimensional constraints): Ed comes from a police family, is focused entirely on his cases and career, and lives in Los Angeles but is completely unfamiliar with the Hollywood world. His partner Jack recognizes the woman as the real star but says nothing --- several earlier scenes have shown the audience that he enjoys manufacturing dramatic moments in everyday life.
\item $H(X)$ (information entropy): The audience has seen the earlier exposition about the prostitution ring and is naturally drawn into Ed's cognitive frame at this point. The surprise of this misidentification is extremely high.
\item $I(X;Y)$ (mutual information): It measures the degree to which this writing choice is ``forced'' by the full-dimensional constraints --- Ed's personality, Jack's motivation, and the previously established prostitution ring setup all converge to make this misidentification almost the only reasonable direction.
\end{itemize}

Jack then whispers to Ed: ``She's the real thing.'' Ed gets a drink thrown in his face by the movie star; Jack laughs in the background. The audience experiences a moment of highly calibrated extreme surprise --- this is one of the film's most memorable passages.

\subsection{The Key Insight: The Weight of the Formula Lies in Conditional Entropy}

The core of understanding this framework is: \textbf{the quantity that actually carries the discriminative function in the formula is conditional entropy $H(X|Y)$, not information entropy $H(X)$.}

$H(X)$ alone has no creative meaning --- gibberish can have extremely high information entropy. Previous traditions in information aesthetics \citep{birkhoff1933aesthetic, moles1968information} mainly worked on the ``complexity'' / ``surprise'' side and were never able to resolve the ``surprise vs. noise'' problem. The subtraction structure of the mutual information formula solves this:

\begin{itemize}
\item When $H(X|Y) \to 0$, feasible choices collapse to very few after full-dimensional constraints are given. At this point $H(X)$ rises automatically --- one of the extremely few solutions forced by the full constraints is, from the unconstrained perspective, exactly a rare choice.
\item High $H(X)$ (non-mediocrity) is the automatic consequence of low $H(X|Y)$ (calibration), not an independent goal.
\item The engine of the whole mechanism is the specificity of $Y$: the richer and more specific the constraints, the lower $H(X|Y)$ is driven, and the higher $I(X;Y)$ rises.
\end{itemize}

\subsubsection{Foundational Property: Conditioning Only Reduces Uncertainty}

The above discussion relies on a foundational property of information theory. We state it explicitly here:

\textbf{Conditioning reduces entropy}:

\[ H(X|Y, Z) \leq H(X|Y) \]

That is: given knowledge of $Y$, introducing any additional valid information $Z$ can only lower conditional entropy or leave it unchanged --- it can never raise it. Equality holds only when $Z$ is conditionally independent of $X$ given $Y$ (i.e., $Z$ carries no new information about $X$).

The special case $H(X|Y) \leq H(X)$ (setting $Z = Y$, starting from the unconditional) also holds.

The implication for creative analysis is direct: \textbf{the richer the constraint dimensions (introducing more components of $Y$), the further the uncertainty of writing choices can only be reduced.} There is no possibility that ``adding too many constraints creates more confusion'' --- every genuine constraint dimension does positive work. This is the mathematical guarantee for the argument chain ``$Y$ more specific → $H(X|Y)$ lower → $I(X;Y)$ higher'' in \S~2.2.

Returning to the \textit{L.A. Confidential} example in \S~2.1, we can see the creative-level implication of this property. There are multiple layers of constraint between detectives Ed and Jack: earliest, during a scandal involving police beating a group of suspects, Ed stood up to report his colleagues --- in Jack's eyes, Ed is someone who advances his career by betraying his partners. Later, while investigating the Night Owl massacre, Ed willingly follows up on one of Jack's leads, and Jack begins to warm to him. Each new constraint layer introduced ($Z$) further reduces the uncertainty $H(X|Y, Z)$ of their interactions --- this is the concrete expression of ``conditioning only reduces uncertainty'' in creative practice.

\subsection{Defining $Y$: The Intersection of Full-Dimensional Constraints}

In standard applications of information theory, $Y$ typically refers to textual context. This paper redefines $Y$ as the intersection of full-dimensional real-world constraints:

\[ Y = Y_{\text{ethos}} \cap Y_{\text{mythos}} \cap Y_{\text{lexis}} \cap Y_{\text{dianoia}} \cap \cdots \]

The four dimensions listed --- ethos (character), mythos (plot), lexis (diction), dianoia (theme) --- are inherited from Aristotle's \textit{Poetics}\footnote{This four-dimensional division corresponds to the first four of Aristotle's six elements of tragedy in \textit{Poetics} Chapter 6 (mythos, ethos, dianoia, lexis, melos, opsis) that apply to written narrative. The last two (melos/opsis, i.e., music and spectacle) belong to theatrical performance and fall outside the scope of prose narrative discussed in this paper. The specific dimension boundaries are not critical to the argument --- the core claim depends on the structural requirement that ``all relevant dimensions are simultaneously satisfied,'' not on the specific count of dimensions (see \S~2.3).}, Chapter 6, which divides tragedy into constituent elements. They apply equally as the most representative constraint categories in prose narrative, but they are not an exhaustive closed list. Different genres may introduce additional dimensions (e.g., meter constraints in poetry, performability constraints in screenplays). The arguments below rely on the structural requirement that ``all relevant dimensions are simultaneously satisfied,'' not on the specific count of dimensions.

\begin{itemize}
\item $Y_{\text{ethos}}$ (character): the psychological and social constraints of characters as human beings in a given situation.
\item $Y_{\text{mythos}}$ (plot): the causal logic among events, plus the internal consistency constraints of the world in which events occur.
\item $Y_{\text{lexis}}$ (diction): the constraints of word choice, rhythm, point of view, and voice in fitting what is being written.
\item $Y_{\text{dianoia}}$ (theme): the constraints of how local choices serve and cohere with the work's overall meaning.
\end{itemize}

The word ``full'' does not mean an exhaustive closed list. It means all relevant dimensions must be satisfied \textbf{simultaneously}. The problem with the well-made play is that it is precise on mythos (plot) but hollow on ethos (character) and dianoia (theme) --- $Y$ is incomplete. Longinus's ``flawless mediocrity'' refers to perfect lexis (diction) but hollow dianoia (theme) --- $Y$ is still incomplete.

\subsection{Three-Way Convergence: The Legitimacy Guarantee for $Y$}

The dimensions of $Y$ are not the researcher's arbitrary list. Their legitimacy is guaranteed by \textbf{three-way convergence}:

\begin{quote}
Author intent × reader's reasonable expectation × the logic of reality → three fully independent sources of judgment simultaneously pointing to the same conclusion.
\end{quote}

The independence of the three sources is the key. When all three give consistent judgment on the same choice, the objectivity of that judgment is guaranteed by multi-source consistency --- analogous to how physics cross-validates the objective existence of a physical quantity through multiple independent experimental methods.

Each dimension of $Y$ has a clear position in the three-way convergence: ethos (character) and mythos (plot) are primarily constrained by the logic of reality (the operating logic of humans as humans, and worlds as worlds); lexis (diction) and dianoia (theme) primarily reflect the internal consistency of author intent; and the three-way convergence receives cross-validation at the reader's reasonable expectation.

Each component of the three-way convergence has an independent scholarly lineage: author intent is the central concern of the hermeneutic tradition \citep{hirsch1967validity}; audience expectation is the subject of reception aesthetics \citep{jauss1982aesthetic, iser1978act}; and the constraint of the logic of reality runs through the mimesis tradition from Aristotle onward. Prior work has also attempted to examine all three together \citep{sell2000literature}, but primarily aimed at linguistic analysis of communicative acts rather than a criterion for judging creative quality. This paper's contribution is to establish three-way convergence as the operational definition of quality judgment and to give it a precise information-theoretic formulation.

In the history of literary criticism, the question of interpretive authority has been long contested. In medieval Europe, the authority to interpret religious texts was monopolized by the literate clergy --- ordinary believers had no right to read or interpret scripture directly. The production of meaning was institutionally locked to a small number of hands. The dismantling of this knowledge monopoly took the long process from the Reformation to the Enlightenment. The pre-modern tradition concentrated interpretive authority in the author: the intentionalism of hermeneutics \citep{hirsch1967validity} holds that the correct meaning of a work is determined by author intent, and readers have no right to establish alternative readings. This position still had strong defenders in the twentieth century --- Vladimir Nabokov insisted that the author's interpretation of a work has ultimate authority and explicitly opposed readers projecting themselves into the text.

In the second half of the twentieth century, the pendulum swung the other way. \citet{barthes1967death} declared ``The Death of the Author,'' announcing that once a work is published the author's monopoly on meaning ends. Reception aesthetics \citep{jauss1982aesthetic, iser1978act} further established the reader's horizon of expectation as a necessary component of literary meaning --- without the reader, meaning cannot be realized. This shift was not purely a theoretical event. Behind it was the broader movement of the non-European world --- after the colonial system began loosening in the twentieth century --- to claim cultural interpretive authority.

Today, people take it for granted that content consumers have every right to interpret and comment on a work. A YouTube user posting a comment beneath a video has no awareness that this ``self-evident'' right they exercise was not self-evident in the Middle Ages. The democratization of interpretive authority --- from clerical monopoly to universal participation --- is the endpoint of this long historical process.

The three-way convergence position of this paper does not take sides with either end of the pendulum. Author intent is one indispensable source of $Y$, but the audience's reasonable expectation has independent legitimacy --- it is not a derivative of author intent but a necessary condition for literary meaning to be realized. The information-theoretic formula offers a neutral mathematical resolution to this centuries-long contest: $Y$ is the intersection of all three; it belongs exclusively to neither the author, the reader, nor reality alone.

\subsection{The Dynamic Dimension: The Chain Rule}

The previous analysis treated $Y$ as statically given. But in real creative writing, $Y$ grows continuously as the narrative unfolds. When the first sentence is written, $Y$ is nearly empty; by the ending, $Y$ is so large that the remaining degrees of freedom are minimal.

The chain rule of information theory captures this dynamic precisely:

\[ I(X_1, X_2, \ldots, X_n ; Y) = \sum_{i=1}^{n} I(X_i ; Y | X_1, \ldots, X_{i-1}) \]

The total information of a novel equals the sum of the information contributed by each choice given all prior choices. Here $X_1, \ldots, X_n$ are ordered by the reader's information-reception sequence (the disclosure sequence), not the order of composition. This rule will be developed further in \S~4.

\textit{L.A. Confidential} again illustrates the colloquial form of the chain rule. Jack recognizing the movie star in the bar but letting Ed embarrass himself is not an isolated event. It is the result of a chain of prior creative decisions. \S~2.2.1 has already shown some constraint dimensions accumulated between Ed and Jack; here we add another: Jack needs to determine whether Ed is a genuinely sincere person or a skilled dissembler --- this matters because it will decide whether to recruit Ed into investigating corruption in the department's upper ranks. Letting Ed embarrass himself is therefore also a test of his character. Under the accumulated constraints of these prior decisions, Jack's behavior is the only natural direction --- $H(X_t | X_{<t}, Y)$ is minimal; this choice is almost the only reasonable one. This is the colloquial version of the chain rule: the ``calibration'' of each creative choice is guaranteed by the accumulated constraints of all choices before it.

\section{Pillar I: The Static Structure --- Constraint Accumulation and Collapse of the Solution Space}

\subsection{The Mathematical Argument for the Core Claim}

This section argues for the mathematical basis of the core claim ``calibrated surprise.'' When $Y$ contains full-dimensional real constraints, $H(X|Y)$ approaching zero and $H(X)$ going high are two faces of the same fact --- the extremely few surviving solutions forced by full constraints are necessarily rare from the unconstrained perspective. The sense of surprise is both what the author consciously pursues as an effect and the automatic mathematical by-product of calibration.

\subsection{The Combinatorial Effect of Constraint Accumulation}

Suppose at some creative decision point, all possible choices form a candidate set $\mathcal{X}$ of size $N$. Suppose $k$ constraint dimensions are independent (or approximately independent), each allowing a fraction $p_i$ to pass. After all full-dimensional constraints are simultaneously imposed, the number of surviving candidates is:

\[ |\mathcal{X}_{\text{full-constrained}}| \approx \left(\prod_{i=1}^{k} p_i\right) \cdot N \]

Even if each dimension excludes only 80\% of choices ($p_i = 0.2$), after four dimensions the fraction satisfying all constraints drops to $0.2^4 = 0.0016$. Only about one or two out of a thousand candidates can simultaneously pass all constraints. This is a thought experiment to illustrate the exponential decay effect, not a literal description of actual creative practice. Real authors do not randomly search candidate space; they use professional experience to navigate directly toward regions satisfying multiple constraints. The key point is: the intersection of full constraints is combinatorially tiny, so choices that fall within it are naturally rare events from the unconstrained perspective.

\begin{figure}[t]
\centering
\includegraphics[width=0.6\linewidth]{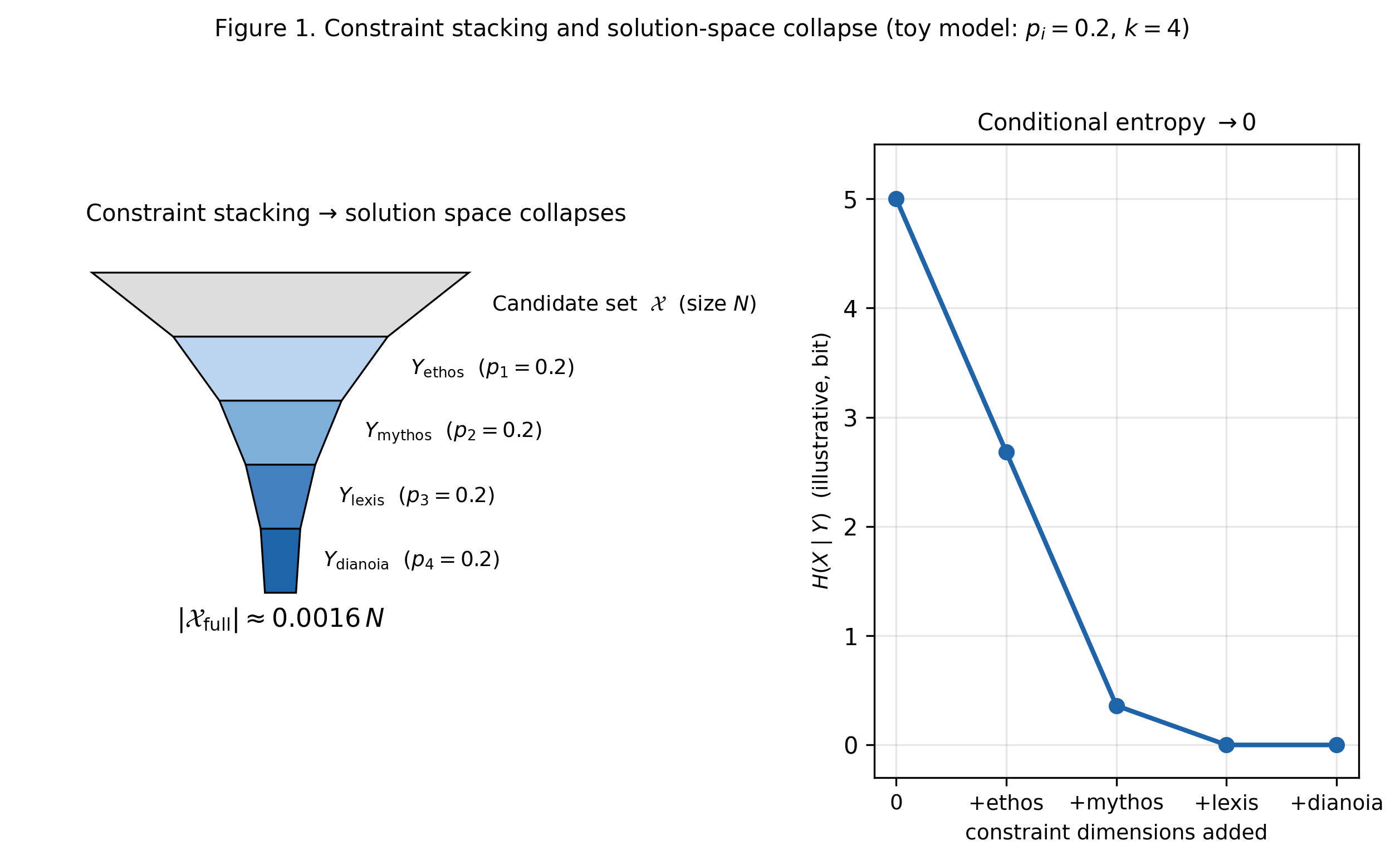}
\caption{Illustration of constraint accumulation and collapse of the solution space. The top layer is the unconstrained candidate set (size $N$); from top to bottom, constraints $Y_{\text{ethos}}, Y_{\text{mythos}}, Y_{\text{lexis}}, Y_{\text{dianoia}}$ are applied in sequence, each with pass fraction $p_i = 0.2$; the bottom layer is the extremely small surviving set $|\mathcal{X}_{\text{full-constrained}}| \approx 0.0016\,N$.}
\label{fig:funnel}
\end{figure}

Translating back into information-theoretic language: when $Y$ is complete, $H(X|Y) \to 0$ (choices are nearly unique), and $H(X)$ rises automatically (that rare surviving solution is, from the unconstrained perspective, an extremely rare choice). The more dimensions there are, the fewer solutions satisfy the full constraints, and the lower the prior probability of those solutions from the unconstrained perspective --- i.e., the less mediocre.

\subsection{Why ``Full'' Is a Necessary Condition}

The precise condition for the mutual exclusion claim is ``full-dimensional calibration.'' Partial calibration does not guarantee non-mediocrity.

\textbf{The well-made play} (la pièce bien faite) achieves high precision on the mythos (plot) dimension but is hollow on ethos (character) and dianoia (theme). An incomplete $Y$ means the collapse of the solution space is insufficient --- under an incomplete $Y$, low $H(X|Y)$ is an illusion.

\textbf{``Flawless mediocrity.''} Longinus, in \textit{On the Sublime}, distinguishes ``flawless mediocrity'' from ``great flaws.'' Rhetorical perfection with hollow intent means $Y$ remains incomplete.

Both counterexamples show precisely why ``full'' is a necessary condition: partial calibration only closes some constraint gates; the solution space does not collapse sufficiently.

\textbf{A unified definition of mediocrity.} Handling both counterexamples together: \textbf{mediocrity is the absence state of full-dimensional chain-calibration} --- one or more constraint dimensions fail to remain continuously engaged; chain accumulation breaks down; creative decisions fall back to the high-probability region of the reader's default distribution. The well-made play has a partial dimension drop (the mythos gear turns; ethos/dianoia disengage). Plain-descriptive prose has constraints that were never truly engaged from the first step (each sentence appears to ``correspond to reality,'' but only touches the shallowest layer of physical correspondence; ethos and dianoia never arrive). Both paths share the same endpoint: insufficient $I(X;Y)$. This differs from ``wrong'' (constraints violated, producing factual hallucination) and ``noise'' (random generation without constraints) --- mediocrity is \textbf{the gravitational state when constraints are insufficient}: without applied force, the system rests at the potential energy minimum, i.e., the mode of the reader's expectation distribution. This yields a compact contrapositive: \textbf{full-dimensional calibration and mediocrity are mutually exclusive}. When $Y$ is fully applied and chain accumulation is effective, the solution space converges to the low-probability region from the unconstrained perspective, making the mediocre state mathematically impossible.

\section{Pillar II: The Dynamic Structure --- The Chain Rule and Incremental Constraint Building}

\subsection{$Y$ Is Dynamically Growing}

\S~3 treated $Y$ as a static given constant. But in real creative writing, $Y$ is not given in advance. It grows continuously as the narrative unfolds. Each writing choice $X_i$ does two things at once: it is governed by all the creative choices established in the prior text (subject to constraints), and it creates new constraints for subsequent text (imposing constraints). $Y$ is a dynamic variable that grows continuously with the information-disclosure process.

\subsection{The Chain Rule}

The precise mathematical counterpart of this dynamic process is the chain rule:

\[ I(X_1, X_2, \ldots, X_n \;;\; Y) = \sum_{i=1}^{n} I\bigl(X_i \;;\; Y \;\big|\; X_1, \ldots, X_{i-1}\bigr) \]

The total information of an entire novel equals the sum of the information contributed by each writing choice, conditioned on all prior text. The ordering of $X_1, \ldots, X_n$ is the reader's information-reception sequence (the disclosure sequence), not the order of composition.

\begin{figure}[t]
\centering
\includegraphics[width=0.6\linewidth]{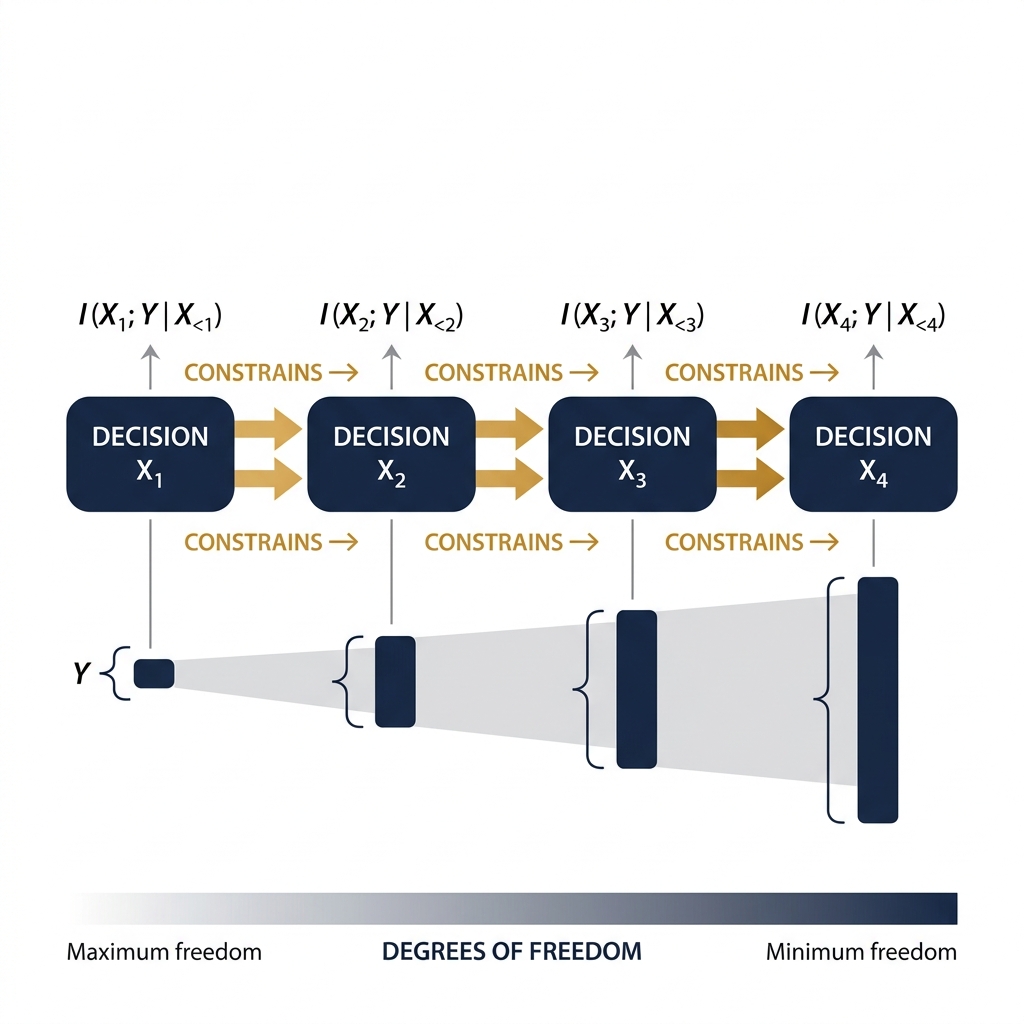}
\caption{The chain rule in creative writing. Each writing choice $X_i$ plays a dual role: it is constrained by all prior choices (receiving constraints) and simultaneously imposes new constraints on subsequent choices (creating constraints). The constraint set $Y$ grows with each decision; the remaining degrees of freedom shrink accordingly. The mutual information contribution $I(X_i; Y | X_{<i})$ at each node measures how much information that choice adds to the chain sum, given everything that came before.}
\label{fig:chain_rule}
\end{figure}

\textit{L.A. Confidential} again. The relationship between Ed and officer Bud is one of fierce antagonism for most of the film, but at approximately the 1 hour 50 minute mark, their relationship transforms into close partnership. This change is far from arbitrary --- a series of specific prior events have been gradually revising each man's assessment of the other. In one interrogation, Ed suddenly uses a bluff to extract a key lead from a suspect --- Bud recognizes that Ed is not just a report-writer but a genuinely intelligent person. Meanwhile, Ed discovers through the case that Bud has real sympathy for victims, does not take bribes, and is loyal --- Ed can no longer dismiss him as the ``mindless thug'' he initially judged him to be. Two detectives who began as fire and water, through each successive creative choice, are forged into close partners by a fistfight in the records room. This transformation is both highly calibrated (each cognitive shift has ample support from prior events) and extremely surprising (audiences at the start of the film would never predict that these two could build trust). The chain rule is visible here: $I(X_t; Y | X_{<t})$ contributes large amounts of mutual information at each cognitive-shift event, and the ``calibration'' of these contributions --- $H(X_t | X_{<t}, Y) \to 0$ --- is entirely guaranteed by all the accumulated creative decisions in the prior text.

\subsection{The Macro/Micro Weighting Problem}

How much weight should macro decisions (story architecture, core setup) and micro execution (a single line of dialogue, a detail) have in quality assessment? This is a problem that has long troubled practice. We once considered designing an ad hoc weighting scheme.

The chain rule reveals that this patch is unnecessary: macro decisions constrain a larger subsequent space and naturally contribute a larger share of information in the chain sum. Micro choices affect the local conditional distribution and naturally receive a smaller share. This is not a design requiring manual calibration; it is a direct mathematical corollary of the chain rule.

\subsection{Methodological Corollary: Dimensions Are Posterior}

The chain rule implies an important methodological principle: dimensions are posterior classifications, not prior frameworks. The chain sum automatically covers all levels of creative choice, with no need to define a dimension checklist in advance and score each item separately.

This principle constitutes a structural critique of current mainstream rubric scoring. Defining a dimension checklist in advance (e.g., ``plot coherence,'' ``character depth,'' ``language style'') and then summing independent scores is essentially a low-resolution approximation of the overall shape of creative quality --- it assumes quality can be orthogonally decomposed, when real creative decisions are simultaneously coupled across multiple dimensions (\S~2.3, \S~3.2). A character's behavioral choice is simultaneously constrained by the intersection of ethos, mythos, and dianoia; scoring it into a single dimension necessarily loses the coupling information between dimensions. The chain rule provides an alternative path that does not depend on prior dimension definitions: let each creative choice be evaluated holistically under its complete prior context; dimensions emerge naturally as post-hoc explanatory labels rather than as a pre-hoc scoring framework.

\section{Core Evidence: Computational Verification}

We use token-level logprobs from an LLM as an operational proxy for the ideal reader's probability distribution. Across 20 pairs (12 Chinese / 8 English) of high-quality vs. systematically degraded literary passages, we verify the core prediction: high-quality passages have systematically higher $I(X;Y)$ than their degraded versions. 20/20 are consistent (\S~5.4). At the end of this section (\S~5.5), we actively declare the proxy boundary between the logprob-based estimator and the theoretical $Y$ --- now that the evidence has been elevated to core status, this construct validity disclosure is an essential part of honest scholarly practice.

\subsection{Experimental Design}

This paper uses logprobs from a large language model as an approximation of the ideal reader's probability judgment, to computationally verify the core prediction of the framework.

The experiment selects 20 pairs of literary passages with known quality differences. Each pair contains an original high-quality version and a version that has been systematically degraded. The degradation method replaces specific, calibrated writing choices with templated, formulaic expressions --- reducing the coupling between writing choices and full-dimensional constraints while keeping the text basically readable.

For each passage, token-level log-probabilities are extracted under two conditions: (1) bare condition (no context), with only the target passage as input; (2) contextualized condition, with the complete prior text as context. From this we compute:

\begin{itemize}
\item $H(X)$: unconditional information entropy --- the surprise of the ideal reader with no background knowledge
\item $H(X|Y)$: conditional entropy with context --- the remaining uncertainty of the ideal reader knowing all prior constraints
\item $I(X;Y) = H(X) - H(X|Y)$: mutual information
\end{itemize}

Core prediction: the $I(X;Y)$ of high-quality passages is systematically higher than the $I(X';Y)$ of their degraded versions.

\subsection{Data Selection}

The experimental corpus contains 20 pairs of literary passages, covering two languages (12 Chinese pairs / 8 English pairs) from 13 different authors. The English corpus is drawn from narrative works by Brandon Sanderson, KJ Parker, Mick Herron, Stephen King, Anthony Horowitz, Jeff VanderMeer, and Tana French. The Chinese corpus is drawn from the literary writing of several contemporary Chinese authors.

Each pair contains: (a) an original high-quality text passage; (b) a degraded version executed by an expert with deep creative experience. Degradation was performed by a team member with more than 20 years of professional writing experience. The operating principle was: within the bounds of keeping the text basically readable and narratively coherent, systematically weaken calibration on full-dimensional constraints --- for example, replacing psychologically authentic character reactions with conventional reactions, or replacing causally tight narrative progression with formulaic turns. This ensures that the degradation is a ``decrease in calibration'' on specific dimensions, not merely word substitution.

\subsection{Implementation Details}

The language model used is the \textbf{Base version} (not Chat/Instruct) of Qwen1.5-7B, deployed on the Alibaba Cloud PAI-EAS platform, with token-level logprobs retrieved via API.\footnote{Reproducibility details: model weights are the public \texttt{Qwen/Qwen1.5-7B} (HuggingFace Hub identifier). The deployment framework is Alibaba Cloud PAI-EAS standard Qwen image. Inference parameters: \texttt{temperature=0}, \texttt{top\_p=1.0}, requesting logprobs only without sampling. The specific deployment instance for the experiments in this paper has been taken offline for cost reasons. Reproducers can reproduce the results by loading the same weights on any inference backend supporting OpenAI-compatible APIs (vLLM, SGLang, TGI, or local transformers). The supplementary materials (appendix) include three parts: (a) complete texts of two representative passage pairs (one Chinese, one English), showing the original and degraded versions in sentence-by-sentence comparison with notes on the degradation operations; (b) the logprobs extraction script for the Alibaba Cloud PAI-EAS API (desensitized, access keys and endpoint URL removed); (c) the mutual information calculation script (including both bare/contextualized request construction, token-level logprobs parsing, and $I(X;Y)$ summary). The complete set of 20 experimental passage pairs is not publicly released, to protect data for subsequent related empirical work; the two sample pairs are sufficient for reproducers to verify that the scripts run and to understand the specific form of the degradation operations.} A 7B-scale general-purpose model was chosen over a larger model or a specially fine-tuned model to verify that the framework can produce effective discrimination even with a medium-scale general model --- if a general 7B model can already systematically distinguish quality differences, we expect larger or more specialized models to provide a useful direction for further verification. The Base version was chosen over Chat/Instruct because Base models have not undergone RLHF or instruction fine-tuning. Their logprobs are closer to the ``intrinsic language model distribution undistorted by preference alignment,'' making them more suitable as approximators of the ideal reader's probability judgment. Chat/Instruct models can show artificial biases introduced by alignment training at certain token positions, making their probability estimates on plain text passages less neutral.

The information entropy calculation formula is:

\[ H = -\frac{1}{n}\sum_{i=1}^{n} \log_2 P(x_i) \]

where $n$ is the token count of the target passage and $P(x_i)$ is the model's predicted probability for the $i$-th token (converted from logprobs). We use the plug-in (per-passage average surprisal) estimator $\widehat{I}(X;Y) = \widehat{H}(X) - \widehat{H}(X|Y)$ as an empirical proxy for the population mutual information; the construct-validity boundary of this proxy is discussed in detail in \S~5.5. The calculation code is provided in the supplementary materials.

\subsection{Results}

Table 1 shows the experimental results for 20 pairs of literary passages.

\begin{table}[t]
\centering
\caption{Mutual information results for 20 literary passage pairs (unit: bit/token)}
\label{tab:results}
\resizebox{\textwidth}{!}{
\begin{tabular}{rllrrrrrrr}
\toprule
\# & Lang & Author & $H(X)$ & $H(X \mid Y)$ & $I(X;Y)$ & $H(X')$ & $H(X' \mid Y)$ & $I(X';Y)$ & $\Delta I$ \\
\midrule
1 & EN & Brandon Sanderson & 4.529 & 3.805 & 0.724 & 3.506 & 3.212 & 0.294 & +0.430 \\
2 & EN & KJ Parker & 3.500 & 3.148 & 0.352 & 3.777 & 3.640 & 0.137 & +0.215 \\
3 & ZH & Author B & 4.526 & 3.435 & 1.091 & 5.304 & 5.076 & 0.228 & +0.864 \\
4 & ZH & Author B & 5.015 & 3.889 & 1.127 & 4.648 & 3.939 & 0.709 & +0.418 \\
5 & ZH & Author B & 4.616 & 3.706 & 0.910 & 4.640 & 4.181 & 0.458 & +0.452 \\
6 & ZH & Author B & 4.494 & 3.641 & 0.853 & 4.747 & 4.296 & 0.451 & +0.402 \\
7 & ZH & Author B & 4.556 & 3.647 & 0.909 & 4.983 & 4.441 & 0.542 & +0.367 \\
8 & ZH & Author B & 5.037 & 3.909 & 1.128 & 5.397 & 4.685 & 0.712 & +0.416 \\
9 & EN & Mick Herron & 4.551 & 4.288 & 0.264 & 4.163 & 4.181 & -0.017 & +0.281 \\
10 & EN & KJ Parker & 3.494 & 3.346 & 0.148 & 4.050 & 3.986 & 0.065 & +0.083 \\
11 & ZH & Jin Cheng & 4.452 & 3.676 & 0.777 & 4.104 & 3.539 & 0.565 & +0.212 \\
12 & ZH & A Yuan & 4.930 & 4.052 & 0.877 & 5.279 & 4.915 & 0.365 & +0.513 \\
13 & ZH & Chun Shen & 4.281 & 3.714 & 0.567 & 4.104 & 3.759 & 0.344 & +0.222 \\
14 & ZH & Hu Yanyun & 4.114 & 3.529 & 0.585 & 3.708 & 3.238 & 0.470 & +0.114 \\
15 & ZH & A Cuo & 3.410 & 2.864 & 0.546 & 3.575 & 3.374 & 0.202 & +0.345 \\
16 & EN & Stephen King & 4.039 & 3.382 & 0.657 & 3.715 & 3.348 & 0.367 & +0.290 \\
17 & EN & Anthony Horowitz & 3.652 & 3.277 & 0.375 & 3.339 & 3.319 & 0.020 & +0.354 \\
18 & EN & Jeff VanderMeer & 5.335 & 4.579 & 0.756 & 4.561 & 4.478 & 0.083 & +0.673 \\
19 & EN & Tana French & 4.257 & 3.450 & 0.808 & 4.046 & 3.748 & 0.298 & +0.509 \\
20 & ZH & Author B & 3.716 & 3.392 & 0.324 & 3.735 & 3.617 & 0.118 & +0.206 \\
\bottomrule
\end{tabular}
}
\end{table}

\begin{table}[t]
\centering
\caption{Summary statistics}
\label{tab:summary}
\small
\begin{tabular}{lr}
\toprule
Metric & Value \\
\midrule
Overall validation rate & 20/20 (100\%) \\
Chinese subset & 12/12 (100\%) \\
English subset & 8/8 (100\%) \\
Mean $I(X;Y)$ (high-quality) & 0.689 bit \\
Mean $I(X';Y)$ (degraded) & 0.320 bit \\
Mean $\Delta I$ & +0.368 bit \\
\bottomrule
\end{tabular}
\end{table}

\begin{figure}[t]
\centering
\includegraphics[width=0.7\linewidth]{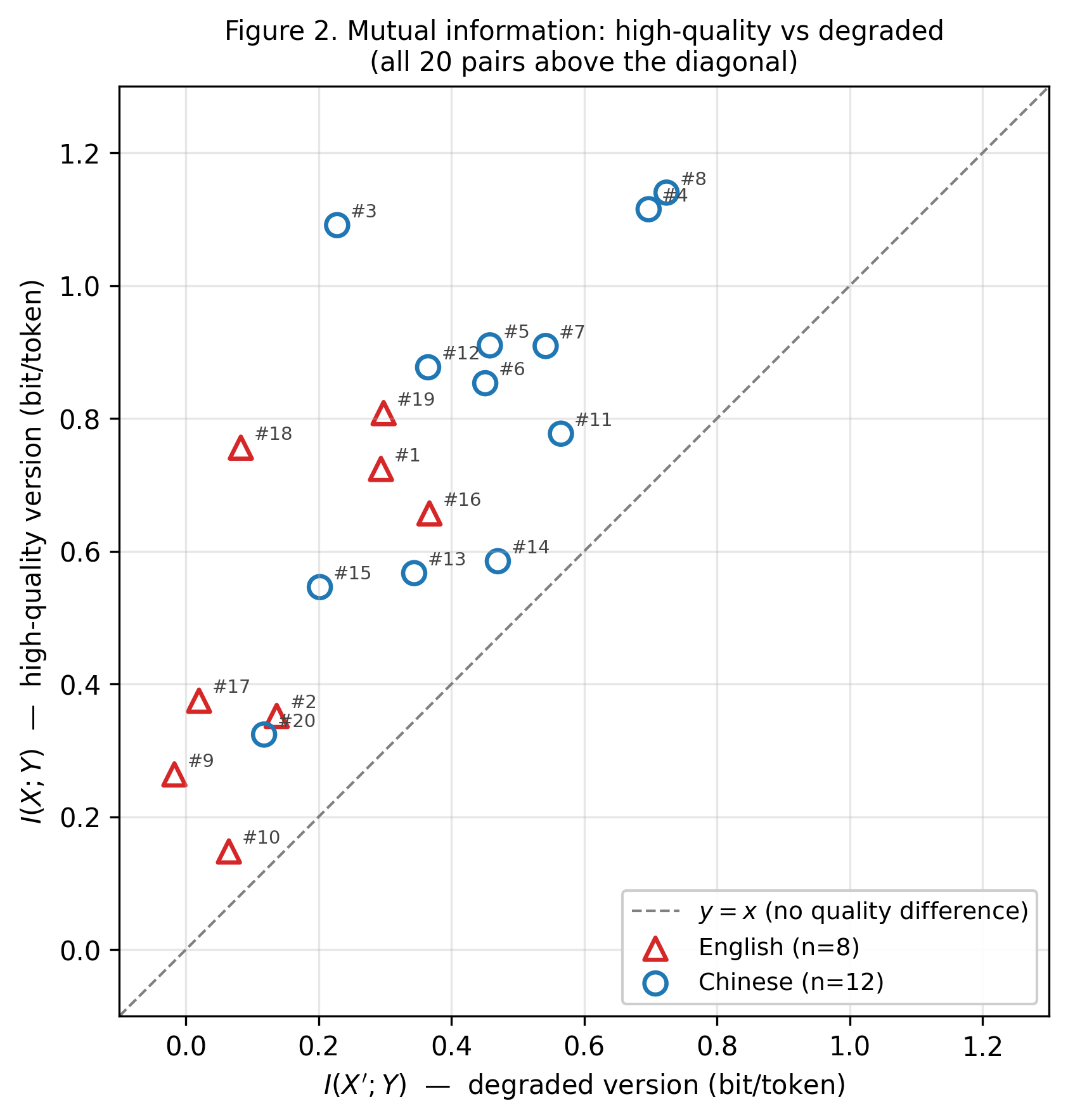}
\caption{Scatter plot of high-quality vs.\ degraded mutual information. Horizontal axis: $I(X';Y)$ (degraded version mutual information, unit: bit/token); vertical axis: $I(X;Y)$ (high-quality version mutual information). 20 data points; Chinese samples (12 pairs) shown as blue circles, English samples (8 pairs) as red triangles; the dashed line is the diagonal $y = x$. All 20 points fall above the diagonal, corresponding to $\Delta I \geq 0$.}
\label{fig:scatter}
\end{figure}

\textbf{Core finding}: All 20 experiments support the core prediction --- high-quality text has systematically higher mutual information $I(X;Y)$ than its degraded version $I(X';Y)$, without exception. This result holds across two languages and works by 13 different authors, showing that the predictive power of the framework does not depend on a specific language or a specific author's style.

It is worth noting that even using a 7B-scale general-purpose language model (rather than a specially trained literary evaluation model), the mutual information differences are systematic and consistent. This is consistent with the theoretical prediction in \S~2.2: the difference in $I(X;Y)$ stems from the information structure of the text itself, not from any special capability of the model. Larger or more refined models are a natural direction for follow-on verification.

\textbf{On the non-monotonicity of $H(X)$.} In Table 1, a small number of pairs (e.g., pairs 2 and 10 by KJ Parker) show $H(X) < H(X')$ --- the unconditional entropy of the high-quality version is actually lower than that of the degraded version. This does not constitute a counterexample to the core theory. The core prediction of this paper is $\Delta I = I(X;Y) - I(X';Y) > 0$, not $\Delta H > 0$. The unconditional entropy $H(X)$ of a single passage is simultaneously affected by local word frequency, sentence length, rare character usage, and other stylistic factors. It therefore need not monotonically reflect quality differences in small samples --- degradation sometimes introduces stylistically ``jarring'' but context-irrelevant low-probability tokens, causing $H(X')$ to rise instead. The decisive metric is the degree to which $H(X|Y)$ drops after context is given: high-quality passages have significantly lower $H(X|Y)$ than $H(X'|Y)$. This is the direct measure of ``context coupling,'' and the true source of the mutual information gap.

\textbf{On negative mutual information.} Pair 9 (Mick Herron, degraded version) shows $I(X';Y) = -0.017 \approx 0$. Note: mutual information is strictly non-negative in theory ($I(X;Y) \geq 0$). The small negative value here comes entirely from estimator variance --- logprobs estimation has statistical fluctuations over finite samples, and small negative values near zero are a normal manifestation of estimation noise, not genuine ``negative mutual information.'' This estimator-level detail does not affect the substantive conclusion: the degraded text has almost no mutual information with the context. This aligns with the theoretical prediction in \S~7.1's relay novel thought experiment that ``text lacking constraint support tends toward zero mutual information,'' and can be seen as natural evidence of the framework's soundness.

\subsection{Construct Validity: The Relationship Between the Proxy and Theoretical $Y$ (Active Disclosure)}

The 20/20 result in \S~5.1–\S~5.4 is encouraging. Precisely delimiting which testable intermediate assumptions its proof strength depends on is part of argumentative completeness --- this section makes those assumptions explicit, not as a weakening of the results.

\textbf{Operational proxy vs. theoretical $Y$}. \S~2 defines $Y$ as ``the intersection of full-dimensional real-world constraints'' --- an idealized object jointly constituted by all relevant dimensions such as ethos, mythos, lexis, and dianoia. In practice, we cannot enumerate this intersection. What actually enters the formula is an operational proxy: \textbf{using the LLM's log-probability for the target token sequence, given vs. not given context, as an approximation of the ideal reader's probability judgment under the corresponding information conditions.} The $I(X;Y)_{\text{proxy}}$ computed in this section is therefore a \textbf{biased lower-bound estimate} of the theoretical $I(X;Y)$: the difference between the LLM distribution and the ideal reader distribution systematically protects the core claim in the degradation-detection direction --- the degree to which the proxy underestimates only makes the true difference larger; the true value can only be stronger.

\textbf{The proxy is directionally consistent with the theoretical $Y$ in most cases}. The LLM's pretraining distribution approximates ``the default linguistic expectation without background knowledge'' (unconditional distribution) and ``the reasonable continuation expectation given context'' (conditional distribution). In most creative scenarios, poor writing that violates full-dimensional constraints $Y$ --- psychologically inauthentic characters, broken causality, misfit diction, incoherent theme --- also shows low coupling on the pretraining distribution: after reading the prior text, the model would not naturally predict such a continuation. So the proxy's high/low is \textbf{directionally consistent} with the theoretical $I(X;Y)$'s high/low in the degradation-detection direction. This is the explainable mechanism behind the 20/20 result in \S~5.4.

\textbf{Expected cases of divergence}. The proxy and the theoretical $Y$ are not strictly equivalent. Expected cases of divergence include at least:
\begin{itemize}
\item (a) An expression is a common word sequence in high-frequency topic corpora (proxy gives low $H(X)$), but is not calibrated under the specific constraints of the current narrative --- proxy underestimates $I(X;Y)$.
\item (b) An accurately rare niche expression is uncommon in the corpus but highly coupled with the context --- the proxy may overestimate $I(X;Y)$ due to high $H(X)$, creating occasional inconsistency with the theoretical direction.
\item (c) The Base model's own corpus biases (e.g., over/under-representation of certain cultural themes, gender issues, specific styles) introduce systematic errors in estimation.
\end{itemize}

\textbf{Delimiting claim strength}. In summary, the 20/20 result in \S~5.4 \textbf{strongly supports} ``the logprob-based estimator can systematically distinguish expert-identifiable degradation from originals.'' The stronger claim that ``the estimator exhaustively captures all forms of calibrated surprise'' is not further claimed by this paper --- that question is left to subsequent work using dimension-by-dimension diagnostic test sets. The theoretical mapping (\S~2–\S~4), the external independent validation (\S~6), and the directionally consistent computational evidence (\S~5) together form the argument chain of this paper. No single support point is sufficient to carry the full claim alone, but their convergence constitutes credible overall evidence.

\textbf{Two visible paths to tightening the proxy} (left to subsequent work): first, replacing the general Base model with a small-scale model fine-tuned with expert chain-of-thought data, directly raising the alignment between the proxy and the theoretical $Y$; second, building a diagnostic test set sensitive to each decision dimension, decomposing proxy error by dimension (ethos/mythos/lexis/dianoia). These are natural follow-on directions; this paper does not develop them here.

\section{External Independent Validation: Literary Traditions Spanning Two Thousand Years Converge on the Same Information-Theoretic Structure}

This section brings together the discussions scattered across \S~3 static unification, \S~4 literary cases, and related work on defamiliarization and classical criticism, and repositions them: they are not the primary subject of this paper's argument, but \textbf{external independent evidence} for the core mapping.

The logical structure is as follows: \textbf{if} mutual information $I(X;Y) = H(X) - H(X|Y)$ truly captures the essential structure of excellent creative writing, \textbf{then} independent literary observations spanning two thousand years, from different traditions, without knowledge of each other, should show non-trivial convergence in this mathematical structure. Below we show that Aristotle's law of probability and necessity (\S~6.1, \S~6.4), Gardner's fictional dream (\S~6.1, \S~6.4), Shklovsky's defamiliarization (\S~6.1, \S~6.3), and Longinus's ``flawless mediocrity'' (\S~6.4) do indeed converge in this way. Two opening examples (\S~6.2) supplement this by showing the concrete form of ``how openings inject constraints into $Y$'' under the chain rule. This constitutes a second type of evidence entirely independent in nature from the computational evidence in \S~5.

\subsection{Three-Way Convergence: Aristotle, Gardner, Shklovsky}

This paper's static framework unifies several scattered concepts in the literary critical tradition within the same coordinate system:

\begin{itemize}
\item \textbf{Aristotle}'s law of probability and necessity --- the classical formulation of the mythos (plot) dimension constraint $Y_{\text{mythos}}$.
\item \textbf{Gardner}'s fictional dream --- maintaining $Y$ across all dimensions simultaneously.
\item \textbf{Shklovsky}'s defamiliarization --- in this framework, defamiliarization is not a goal but a by-product: choices that are full-dimensionally calibrated naturally fall in the tail of the probability distribution.
\end{itemize}

Each of these scholars touched on a different facet of the same mathematical structure from a different angle. The mutual information formula provides the coordinate system for unifying these insights scattered over two thousand years.

\subsection{Literary Cases: How Openings Inject Constraints into $Y$}

\textbf{García Márquez's opening of \textit{One Hundred Years of Solitude}}\footnote{The English quotation follows Gregory Rabassa's classic translation, \textit{One Hundred Years of Solitude} (Harper \& Row, 1970), translated from the Spanish original \textit{Cien años de soledad} (1967).}: ``Many years later, as he faced the firing squad, Colonel Aureliano Buendía was to remember that distant afternoon when his father took him to discover ice.'' $X_1$ itself has extremely high information content (three temporal layers overlapping) and simultaneously injects extremely strong constraints into $Y$ --- locking down the entire novel's tone, temporal structure, and narrative perspective. In the chain sum, this $X_1$ changes the entire conditional distribution of the hundreds of thousands of characters that follow.

\textbf{The opening of \textit{Xiaohuadan} (\begin{CJK*}{UTF8}{gbsn}小花旦\end{CJK*}) by Wang Zhanhei}\footnote{\textit{Xiaohuadan} (\begin{CJK*}{UTF8}{gbsn}小花旦\end{CJK*}) by Wang Zhanhei, a contemporary Chinese author. The English translation of the quoted passage is the authors' own working translation.}: ``Twelve and a half yuan --- that is the distance from Shanghai to my home. If people measured train trips by mileage the way they do flights, mine wouldn't be worth mentioning. One hour. While the people going far away are still queuing to make instant noodles, I've already arrived.'' With a high-speed rail ticket price, a few lines of reflection on ``mileage,'' and the detail of instant noodles, this injects multiple constraints into $Y$: the tension between physical and psychological distance (``not worth mentioning'' implies the narrator's complicated attitude toward home), the narrator's position relative to ``the people going far away'' (self-excluded from those going far), and a tone of cool, restrained detachment. What $X_1$ changes is not just the conditional distribution of subsequent sentences, but the reader's entire framework for judging all character actions thereafter --- a narrative perspective of low-intensity estrangement toward home is locked in.

\textbf{Mediocre opening} (a counterexample constructed for contrast, not attributed to any work): ``The weather was clear, not a cloud in the sky.'' $H(X_1)$ is extremely low (predictable), and this sentence almost generates no constraints on the subsequent narrative --- $Y$ does not grow; it contributes nearly zero information to the chain sum.

\subsection{Correspondence with Cognitive Poetics / the Defamiliarization Tradition}

Shklovsky \citep{shklovsky1917art} held that the function of literature is to break habituated perception. Miall \& Kuiken \citep{miall1994foregrounding} developed foregrounding theory. In this framework, defamiliarization corresponds to $H(X)$ going high. But this paper's advance is: defamiliarization is not the goal; it is the automatic by-product of full-dimensional calibration. Defamiliarization unsupported by $Y$ has high $H(X|Y)$ as well, and mutual information approaches zero.

\subsection{Correspondence with the Classical Critical Tradition of Calibration}

Aristotle \citep{aristotle_poetics} requires stories to follow the law of probability or necessity. Gardner's \citep{gardner1983art} fictional dream requires full-dimensional maintenance without breaking. McKee's \citep{mckee1997story} X-ray machine corresponds to the drop in $I(X;Y)$ when any $Y_i$ is uncalibrated. Longinus's \citep{longinus_sublime} ``flawless mediocrity'' corresponds to low mutual information when $Y$ is incomplete. This paper unifies these discussions scattered over two thousand years in the same information-theoretic coordinate system.

\subsection{Summary: Three Groups of Convergence Form a Predictive Validation}

The three correspondences in \S~6.1–\S~6.4 --- Aristotle's law of probability and necessity $\leftrightarrow$ $H(X|Y) \to 0$; Gardner's fictional dream $\leftrightarrow$ completeness of $Y$; Shklovsky's defamiliarization $\leftrightarrow$ $H(X)$ going high --- are not post-hoc analogies. They are precise structural mappings: each maps an independent scholar's natural-language summary of a creative observation to a specific mathematical position in the mutual information formula $I(X;Y) = H(X) - H(X|Y)$.

If there were only one such correspondence, we could still suspect a rhetorical game of ``forcing modern concepts onto the ancients.'' But \textbf{three independent sources} --- from fourth-century-BC Greece, early-twentieth-century Russia, and late-twentieth-century America --- each take their position at \textbf{different locations} in the same mathematical structure. These positions were not reserved for them when the formula was designed. This convergence indicates that the mathematical structure most likely captures some true underlying regularity, not a rhetorical construction by the author.

The external validation in this section and the computational evidence in \S~5 are entirely independent in nature: the former is a conceptual-level test of historical observations' positioning; the latter is a numerical-level statistical test of contemporary LLM behavior. Their directional consistency jointly supports the $I(X;Y)$ mapping proposed in \S~2.

\section{Sanity Checks and the Ideal Reader}

This section brings together two supporting arguments: the relay novel thought experiment (\S~7.1) --- showing why the subtraction structure of mutual information is naturally immune to ``unconstrained surprise''; and the ideal reader hypothesis (\S~7.2) --- explaining how the probability distribution $P(x)$ in the formula is anchored. Both play supporting roles in the main argument (\S~2–\S~5) and need only one section each; they are therefore combined here.

\subsection{The Relay Novel Thought Experiment}

\subsubsection{The Challenge}

A natural challenge: does high mutual information simply mean ``writing wildly''? Gibberish can have extremely high information entropy.

\subsubsection{The Thought Experiment}

Imagine a ``novel'': two characters sit at a table and decide to play word-association. The entire remainder of the work consists of the two taking turns saying words --- red-braised pork, Venezuela, Spanish, watermelon, clutch... This work has the shell of narrative (characters, setting, action) and is formally a complete novel. Each word jump is extremely surprising, with high $H(X)$.

But the mutual information formula's verdict is: under the rules of ``word association,'' the previous word places no narrative-level constraint on the next word. ``Red-braised pork'' does not require ``Venezuela'' to hold on any dimension of ethos, mythos, lexis, or dianoia. $H(X|Y)$ is near its maximum.

\[ I(X;Y) = H(X) - H(X|Y) = \text{high} - \text{high} \approx 0 \]

Surprising but unsupported by constraints = pure noise = zero mutual information. The formula correctly gives a score of zero.

\subsubsection{Its Role in the Paper}

The relay novel thought experiment does not constitute a ``paradox'' --- the formula gives the correct verdict. Its function in this paper is to show in advance why the subtraction structure of mutual information is naturally immune to ``unconstrained surprise.'' Even if a work has the formal shell of narrative, if its choices are not governed by full-dimensional constraints, $I(X;Y)$ still approaches zero. This is the fundamental reason this paper selects mutual information $I(X;Y)$ rather than information entropy $H(X)$ as the core measure, and the fundamental point of divergence from the earlier tradition of information aesthetics \citep{birkhoff1933aesthetic, moles1968information}.

\subsection{The Ideal Reader Hypothesis}

\subsubsection{The Problem}

Whose perspective does the probability distribution $P(x)$ in the formula reflect? The same sentence produces different information content for different readers. If the output of the formula depends on ``who is reading,'' the objectivity of the framework collapses.

\subsubsection{The Ideal Reader}

The solution is to introduce the ideal reader: the probability distribution in the formula corresponds to ``a hypothetical receiver who theoretically possesses complete aesthetic decoding ability.'' This does not pretend such a person actually exists --- physics does not pretend frictionless surfaces exist either. The ideal reader is a definitional tool for the measurement standard, ensuring that $I(X;Y)$ is an intrinsic property of the work rather than a subjective reaction of a particular reader.

A mediocre reader failing to extract the information content from a passage does not mean the information content is not there --- it only means the reader lacks the decoder.

\subsubsection{Information-Theoretic Correspondence}

In information-theoretic terminology: the work is the source, the reader is the channel. The ideal reader corresponds to a noiseless channel --- one that completely transmits all information from the source without introducing receiver-end distortion. We care about the source-end information (creative quality), so we need to remove the channel-end bottleneck.

\section{Discussion and Limitations}

\S~1.2 and \S~5 have already stated the core implications of this paper for LLM alignment. This section discusses several directions extending from the main argument and the explicit boundaries of this paper. It does not repeat the alignment roadmap itself.

\subsection{The Reader Decoder: From Measuring Works to Measuring Audiences}

The same formula can be used in reverse: fix the work and vary the reader side, measuring reception differences across different audience groups for the same work. Let $P_{\text{expert}}(x)$ be the probability distribution of a deep reader and $P_{\text{casual}}(x)$ that of a casual reader. The taste gap between the two groups can be quantified with KL divergence:

\[ D_{\text{KL}}(P_{\text{expert}} \| P_{\text{casual}}) \]

This paper only offers a forward-looking note, marking two potential directions: first, an information-theoretic explanation of ``works ahead of their time'' in literary reception history (what changes is not the work but the average social audience's $Y$, which has thickened over time); second, the quantification of the information gap between expert chain-of-thought and crowdsourced chain-of-thought in AI alignment, providing a theoretical standard for subsequent CQA methodology papers.

This reverse-applied framework for audience modeling has a specific methodological implication. When we try to adjust the probability distribution in the formula from the ideal reader to a specific audience group (e.g., the casual readers represented by $P_{\text{casual}}(x)$), a natural but wrong approach is to survey that group at scale and ask them to self-report their aesthetic preferences. The fundamental difficulty of this approach is: an audience's tacit aesthetic judgment is procedural knowledge, and they cannot accurately articulate in natural language why they find a piece of writing good or bad. What they can offer is at most a post-hoc rationalization, with a systematic gap from the true judgment mechanism.

A more reliable approach is to find an expert who simultaneously possesses two capabilities: first, they have deep aesthetic decoding ability themselves, and can therefore clearly articulate the content and weight of each constraint dimension $Y_i$; second, they have a highly accurate mental model of the target audience group and can accurately predict that group's conditional probability distribution $P_{\text{casual}}(x \mid y)$ at specific decision points. Such a person is far rarer than an ``ordinary reader,'' but the signal-to-noise ratio of a single expert's signal is far higher than the statistical average from large-scale surveys.

This principle is epistemologically isomorphic with the ``right person over right process'' approach advocated in one of our related works: when it comes to acquiring tacit knowledge, \textbf{finding one expert with complete introspective and modeling ability is always better than large-scale sampling of a group that cannot self-articulate.} This is the same methodological position surfacing twice in different problems.

\subsection{Connection to Subsequent Papers}

This paper defines \textbf{what creative quality is} (mathematical definition). The overall research plan unfolds across four complementary levels; the other three are natural follow-on directions: a method for eliciting expert chain-of-thought data, addressing ``how to obtain the training data needed''; an empirical study fine-tuning a small-scale LLM with approximately 100 expert chain-of-thought examples, testing whether $H(X|Y)$ decreases systematically after fine-tuning; and a diagnostic test set targeting micro-level narrative decision points, providing dimension-by-dimension calibration diagnostics independent of training data via logprobs. Together these form a complete chain: ``theoretical definition → data production → empirical loop → independent diagnostics.''

\subsection{Theoretical Extension: The Humor Mechanism as a Representative Case of Isomorphism}

\S~1.2 already noted that ``calibrated surprise'' as an information-theoretic structure can take concrete forms at different time scales and different levels of abstraction, and listed four examples: a humorous punchline, a narrative reversal, Shklovsky's defamiliarization, and \textit{The Godfather}'s wholesale reframing of the ``criminal'' category. This section takes \textbf{humor} as the representative case for full development, for two reasons. First, humor is the subtype among the four examples with the shortest time scale and the clearest boundaries between $Y$ and $X$; the mathematical mapping can be completed in the shortest space. Second, cognitive linguistics has already provided a ready-made, widely accepted explanatory framework for humor (Incongruity-Resolution Theory), which enables a rigorous mathematical isomorphism comparison. The development of the other three examples is structurally parallel, differing only in the specific composition of $Y$ and the time scale (narrative reversals have a $Y$ spanning thousands of words of causality and character setup; defamiliarization's $Y$ spans the reader's conventionalized perceptual expectations for the object described; category reframing's $Y$ spans the reader's prior distribution over an entire category). This section therefore does not elaborate on each individually.

In detail: Incongruity-Resolution Theory --- the dominant framework in cognitive linguistics for explaining humor --- exhibits a perfect mathematical isomorphism in this model. A high-quality punchline is extremely anomalous without context ($H(X)$ very high), yet under the intersection of constraints formed by the preceding setup and the listener's implicit cultural presuppositions ($Y$), it becomes the single perfect solution ($H(X|Y) \to 0$), triggering extremely high mutual information $I(X;Y)$. In this sense, humor is not a special case separate from narrative fiction, but the extreme compression of the same information-theoretic mechanism into an extremely short time scale.

The isomorphism of humor therefore provides double support for the core claim. On one hand, it is itself a complete instantiation of ``calibrated surprise.'' On the other hand, it supports the stronger claim proposed in \S~1.2 --- that the four examples (and broader micro-pragmatic phenomena) share the same information-theoretic structure. The framework has broad explanatory reach across genres and scales.

\subsection{Extension: From Creative Surprise to the Surprise Profile of Discursive Truthfulness}

One extension of this paper's framework worth recording is the potential to extend the diagnostic gaze of ``calibrated surprise'' from fictional creative writing to truthfulness detection in non-fictional discourse --- particularly the honesty/deception detection subfield that has been active in alignment research in recent years.

Core conjecture: \textbf{fabricated discourse, because it is not governed by full-dimensional constraints of reality $Y$, will have a surprise temporal distribution that is systematically more uniform; the surprise profile of truthful discourse, however, must be ``fast and slow with structure''} --- at fact nodes constrained by real $Y$, surprise naturally is lower (because these nodes are almost the only solutions forced by reality $Y$); at reasonable elaboration nodes, surprise is higher (because the speaker introduces new information).

Formally, this gives a \textbf{falsifiable prediction}: not detecting content truth or falsity, but detecting the temporal distribution characteristics of $-\log P(x_t | x_{<t})$ in discourse sequence $\{x_t\}$ --- fabricated discourse should show a smoother, more uniform surprise curve, because the speaker lacks real $Y$ to constrain the rhythm externally.

Potential applications include: public earnings call statements by listed company CEOs, political speeches, and post-hoc truthfulness filtering of AI-generated content. This paper does not conduct experimental verification here but records this implication as a follow-on direction --- it connects the core structure of this paper (the coupling of full-dimensional constraints $Y$ and surprise) to the adjacent but independent research trajectory of honesty assessment in alignment.

\subsection{Limitations}

\begin{enumerate}
\item \textbf{Formalization of dimensions}: This paper follows the \textit{Poetics} tradition in dividing the core constraints of $Y$ into ethos (character), mythos (plot), lexis (diction), and dianoia (theme). This division serves argumentative clarity, not an exhaustive enumeration of all relevant constraints --- different genres may introduce additional dimensions (see \S~2.3). The dimensions are coupled in real creative writing (e.g., character psychology and event causality often shape each other), so \S~3.2 explicitly marks the combinatorial estimate as ``a thought experiment, not a literal description of actual creative practice.'' More fine-grained dimension partitioning and empirical characterization of coupling structure are left for subsequent work.
\item \textbf{Operationalization of the ideal reader}: LLM logprobs as an approximation of the ideal reader is a viable direction, but the LLM's own biases will introduce systematic errors.
\item \textbf{Granularity of chain structure}: The granularity definition of ``one writing choice'' in the chain sum and its effects on results need further investigation.
\item \textbf{Cross-cultural applicability}: The examples in this paper come primarily from Chinese and Western classical traditions. Whether the framework applies equally to other cultural traditions remains to be verified.
\end{enumerate}

\section{Related Work}

\subsection{Information Aesthetics}

Birkhoff (1933) proposed the aesthetic measure ``order/complexity.'' Moles (1968) used Shannon entropy to analyze the information content of artworks. \citet{ramachandran1999science} proposed laws of aesthetic experience from a neuroscientific perspective.

The key distinction between this paper and the above work: we use mutual information $I(X;Y)$, not information entropy $H(X)$ or an order-complexity ratio. Pure surprise cannot distinguish ``meaningful surprise'' from ``noisy surprise.'' Introducing the constraint side $Y$ makes this distinction possible.

\subsection{Cognitive Poetics}

Shklovsky's \citep{shklovsky1917art} defamiliarization theory and Miall \& Kuiken's \citep{miall1994foregrounding} foregrounding theory and their relationship with this framework are developed in \S~6.3. They are not repeated here; citation entries are retained.

\subsection{The Calibration Tradition in Literary Criticism}

The correspondence between this framework and the classical and modern discussions of creative calibration in Aristotle \citep{aristotle_poetics}, Gardner \citep{gardner1983art}, McKee \citep{mckee1997story}, and Longinus \citep{longinus_sublime} is developed in \S~6.1 and \S~6.4. It is not repeated here.

\subsection{Kolmogorov Complexity}

Kolmogorov complexity \citep{kolmogorov1963tables} measures the length of the shortest program needed to describe an object. The core distinction from this paper's use of mutual information is: $I(X;Y)$ explicitly includes the constraint side $Y$, measuring the information transmitted between choices and constraints, not the information needed to describe the object itself. The analysis of creative quality inherently requires a reference frame (constraints); mutual information is therefore the more appropriate tool.

\section{Conclusion}

This paper provides a computable theoretical anchor for LLM Creative Quality Alignment (CQA): mutual information $I(X;Y) = H(X) - H(X|Y)$.

When a writing choice simultaneously satisfies full-dimensional constraints $Y$, feasible solutions are forced into an extremely narrow space ($H(X|Y) \to 0$, ``calibrated''); these surviving solutions are naturally low-probability from the unconstrained perspective ($H(X)$ goes high, ``surprising''). Calibration and non-mediocrity are not two goals to pursue separately. They are two sides of the same set of constraints --- full-dimensional creative calibration and mediocrity are mathematically mutually exclusive.

This paper provides two types of evidence entirely independent in nature for this mapping. \textbf{Computational evidence} (\S~5): using LLM logprobs as an operational proxy to compute $I(X;Y)$ on 20 pairs (Chinese/English) of high-quality vs. systematically degraded literary passages, with 20/20 supporting the core prediction; \S~5.5 actively declares the proxy boundary. \textbf{External independent validation} (\S~6): Aristotle's law of probability and necessity, Gardner's fictional dream, Shklovsky's defamiliarization --- literary critical observations scattered over two thousand years from mutually independent traditions --- each take their position at different locations within the same mathematical structure. The directional consistency of the two types of evidence jointly constitutes the overall argument strength of this paper.

The two theoretical pillars (\S~3–\S~4): \textbf{static} --- the combinatorial accumulation of full-dimensional constraints leads to collapse of the solution space; $H(X|Y) \to 0$ automatically produces high $H(X)$. \textbf{Dynamic} --- the chain rule ensures that each writing choice is both constrained by prior choices and imposes constraints on subsequent ones; macro-level decisions naturally obtain a larger share of total information, with no ad hoc weighting.

The direct implication for engineering practice: improving LLM creative quality judgment is equivalent to calibrating the internal conditional probability distribution $P(x|y)$ to systematically reduce $H(X|Y)$. Specific methodology and evaluation benchmarks are natural follow-on directions based on this anchor. This paper does not undertake them but provides their theoretical foundation.

In a word: \textbf{the essence of excellent creative writing is calibrated surprise; for LLMs, the pursuit of calibrated surprise is equivalent to maximizing $I(X;Y)$ under full-dimensional constraints $Y$}.

\nocite{shannon1948mathematical, anonymous2026suppressed}

\bibliographystyle{tmlr}
\bibliography{references}

\appendix
\section{Supplementary Material: Two Representative Sample Pairs and Notes on the Degradation Work}
\label{app:samples}

This appendix gives two representative sample pairs from the experiment in \S~5---one in Chinese (Table~\ref{tab:results}, row 12, A Yuan, \emph{Mirror-Flowers}) and one in English (Table~\ref{tab:results}, row 16, Stephen King, \emph{End of Watch})---with the full $X$ text shown side by side across the high-quality and degraded versions, and a short note on what the degradation work looked like in practice. The prior context $Y$ for the two pairs (about 22{,}000 Chinese characters and 70{,}000 English characters respectively) is not reproduced here, for copyright reasons; only a short scene summary is given. The full 20-pair experimental corpus, together with the $Y$ context and the token-level logprob arrays, is not publicly released, out of consideration for our follow-up CQA work.

\subsection{Sample pair \#16: Stephen King, \emph{End of Watch} (English)}

\paragraph{$Y$ prior context (scene summary).}
$Y$ is the entire narrative up to the start of Chapter~6 of King's \emph{End of Watch} (Scribner, 2016). The plot involves the national speaking tour of feminist speaker Kate McKay and her assistant Corrie---who, in a previous chapter, was attacked by a protester at a street event, with her skin burned by pepper spray and her eyes red. The closing passage of $Y$ records officer Hughes finishing taking a statement on the scene, with an appointment to repeat the case at the Second Street precinct at 1:30 the next day. The $X$ in this section continues right after that exchange.

\paragraph{$X$, high-quality version (original).}
\begin{quote}
\begin{flushleft}\itshape
6\par\medskip
When Hughes is gone, Kate says, ``I want you onstage tonight. Are you good with that?''\par\medskip
Corrie feels a bolt of fright at the idea. ``Would I have to speak?''\par\medskip
``Not if you don't want to.''\par\medskip
``Then okay. I guess.''\par\medskip
``You don't mind being Kate McKay's object lesson? Don't resent me for it?''\par\medskip
``No.'' Is that the truth? Corrie wants it to be.\par\medskip
``I want to take your picture. While your eyes are still red and puffy and your skin is still irritated. All right?''\par\medskip
``Yes.''\par\medskip
``People need to understand there's a price for standing up. But it can be paid. They need to understand that, too.''\par\medskip
``Okay.''\par\medskip
\textit{I've become a selling point}, Corrie thinks. She sees Kate's willingness to do this, to seize this, as a character flaw, but also as a character strength. That it can be both is a new idea for her.
\end{flushleft}
\end{quote}

\paragraph{$X'$, degraded version (produced by an experienced writer).}
\begin{quote}
\begin{flushleft}\itshape
6\par\medskip
When Hughes left, Kate said, ``I want you to sit on the stage tonight. Is that okay with you?''\par\medskip
Corrie felt a little bit tired at the thought of it. ``Will I have to say anything?''\par\medskip
``No, not if you don't feel like it.''\par\medskip
``Then that's fine, I guess.''\par\medskip
``You don't mind helping me out with the presentation? It's not a problem?''\par\medskip
``No.'' Corrie thought that it was probably true.\par\medskip
``I'd like to take a photo of you. Even though you look a bit worn out from the day. Is that alright?''\par\medskip
``Yes.''\par\medskip
``People just need to see who is involved. It's part of the process. They should understand that.''\par\medskip
``Okay.''\par\medskip
\textit{I'm just part of the plan}, Corrie thought. She saw that Kate was simply being practical and trying to get things organized. It seemed like a very normal way for Kate to handle the situation.
\end{flushleft}
\end{quote}

\paragraph{Notes on the degradation work.}
The degradation here was carried out by feel, not by a pre-set plan to hit specific dimensions; no plot move is changed and no grammatical errors are introduced. Reading after the fact, the differences from the original fall mainly on two dimensions, ethos and dianoia:
\begin{itemize}
\item \textbf{Corrie's inner reactions become generic.} The original ``a bolt of fright'' (her reflexive fear of going onstage), the mid-passage rhetorical question ``Is that the truth? Corrie wants it to be.'' (the slide into self-doubt), and the sharp self-naming at the end---``\textit{I've become a selling point}''---are replaced in $X'$ by ``a little bit tired'', ``probably true'', ``\textit{I'm just part of the plan}'', i.e.\ flat, neutral self-reports with no temperature and no contradiction. The precise grain of the character's inner state ($Y_{\text{ethos}}$) is flattened.
\item \textbf{The ethical tension in Kate's proposal is removed.} The original ``object lesson'', ``Don't resent me'', ``there's a price for standing up''---each line pushes towards the moral grey area of ``using a victim to drive mobilisation''. In $X'$ these are replaced by procedural phrases such as ``helping me out with the presentation'' and ``part of the process'', and the local thematic load ($Y_{\text{dianoia}}$) carried by the character's ethical bind is dissolved into a routine organisational matter.
\item \textbf{The cognitive break of ``two sides at once'' at the end is replaced by a linear rationalisation.} The original lets Corrie identify, on the same object, both ``character flaw'' and ``character strength'', and explicitly marks ``That it can be both is a new idea for her''---a King-style micro-structure of cognitive turn. $X'$ replaces it with ``simply being practical'', ``a very normal way to handle the situation'': the turn is flattened into ``this is just a normal way to handle things''.
\end{itemize}

After the degradation, $X'$ is still smooth at the level of vocabulary and syntax (the lexis dimension is largely preserved), but its coupling with $Y$ on ethos and dianoia drops sharply. $H(X' \mid Y)$ rises about $0.03$ bit/token over $H(X \mid Y)$, giving $\Delta I = +0.290$ bit/token (see Table~\ref{tab:results}, row 16).

\subsection{Sample pair \#12: A Yuan, \emph{Mirror-Flowers} (Chinese; English translation)}

\paragraph{$Y$ prior context (scene summary).}
$Y$ is the first half of a contemporary Chinese short story, \emph{Mirror-Flowers}. The narrator ``I'' (``Zhuzhu'') is a teacher in the Chinese department at a university and has a long-running afternoon-tea relationship with a woman friend ``Yan Li''. Yan Li keeps describing, in front of the narrator and in a boastful tone, the intimate details of her life with a boyfriend named ``Shen''---they watch films together, stay at hot-spring inns, attend a lecture on \emph{Methods of Composition}, and so on. After one Tuesday at the cinema, the narrator gradually realises that Yan Li sat alone in the theatre for three hours, with no ``Shen'' beside her. The end of $Y$ leads into another afternoon tea: Yan Li has just brought up a rumour that the small housekeeper of another woman friend, Su Beiyan, was slapped, and uses it to launch a new round of ``Shen'' stories. The $X$ in this section continues right after that.

\paragraph{$X$, high-quality version (translated from the Chinese original).}
\begin{quote}
\begin{flushleft}\itshape
This Tuesday afternoon, Shen and I went to ``Skyscreen'' to see a film.\par\medskip
Ann Hui's \emph{The Golden Era}.\par\medskip
This Shen, sometimes he is really annoying. The film was three hours long, and for all three hours he kept holding my hand, sweaty and sticky, and he didn't even mind the heat.\par\medskip
And he whispered in my ear: this is what they mean by ``hand in hand, growing old together''.\par\medskip
After the film we went for congee, at \emph{Floating Life}---do you know \emph{Floating Life}? In the north of the city, a newly opened congee place, the side dishes there are very fine. The shredded winter melon is as green as jade; the sweet-rice-wine kidney-beans with scallion are also good, soft and sweet. Zhuzhu, why don't we go to \emph{Floating Life} for congee today?\par\medskip
This was really strange. That Tuesday I clearly saw that Yan Li was alone, alone from start to finish, sitting bolt upright by herself, leaving the theatre alone in something like a flustered way. Where was the ``hand in hand'' Shen by her side?\par\medskip
Could it be that Shen is a ghost? One that only Yan Li can see, and no one else?\par\medskip
But are there ghosts in this world?\par\medskip
And then it suddenly came to me: maybe there has never been any Shen at all. The lecture on \emph{Methods of Composition}, the blue-grey bathrobe at the hot-spring inn, the whisper of ``hand in hand, growing old together''---all of it is no more than Yan Li's daydream! A daydream, no more!\par\medskip
A wave of sadness came over me.\par\medskip
And there she was, still talking, her eyes shining, her face like a peach blossom, like a young female lead on an old opera stage.
\end{flushleft}
\end{quote}

\paragraph{$X'$, degraded version.}
\begin{quote}
\begin{flushleft}\itshape
This Tuesday afternoon, Shen and I went to ``Skyscreen'' to see a film.\par\medskip
We saw a long drama film.\par\medskip
This Shen is pretty quiet during a film. For the three hours we just sat there in our seats and watched, and we didn't talk much.\par\medskip
At one point he just leaned over and said, this film is pretty long.\par\medskip
After the film we went for congee, at an ordinary congee place in the north of the city. The side dishes there were just the usual home-cooked kind. There was a plate of stir-fried winter melon and a portion of kidney-beans, the taste was nothing special, just something to eat. Zhuzhu, why don't we go there for a casual meal today too?\par\medskip
I remembered that on Tuesday I did see Yan Li. She had just come out of the screening hall by herself. Probably Shen had gone to the bathroom or was in line buying water.\par\medskip
Even though Shen wasn't beside her at that moment, when two people watch a film and have a meal together, splitting up for a bit in the middle is a very normal thing.\par\medskip
A weekend usually goes more or less like that for everyone, by the book. So I didn't think anything more of it.\par\medskip
And there she was, still talking slowly, her face calm, her tone flat, as if she were just chatting about some very ordinary little daily thing.
\end{flushleft}
\end{quote}

\paragraph{Notes on the degradation work.}
The degradation here was again carried out by feel, with the plot moves preserved (``Tuesday at the cinema + congee afterwards + Yan Li keeps talking'') and no grammatical defects introduced; reading after the fact, the differences from the original fall on three dimensions: ethos, lexis and dianoia.
\begin{itemize}
\item \textbf{The ``over-specific'' boasting texture in Yan Li's narration is rubbed flat---this is the most visible difference after the fact.} In the original, Yan Li's ``Shen'' narration has a clear boasting texture: a named director and film (``Ann Hui's \emph{The Golden Era}''), a named congee place (``\emph{Floating Life}''---newly opened in the north of the city, with fine side dishes), two named dishes (``shredded winter melon as green as jade'', ``sweet-rice-wine kidney-beans, soft and sweet''), quotation-level intimate markers (``sweaty and sticky'', ``hand in hand, growing old together''), and even a reverse rhetorical move (``sometimes he is really annoying''---boasting intimacy in the form of a complaint). In $Y$, these over-specific details are exactly what lets the narrator identify the ``daydream''---only what did not happen needs to be coloured in over and over. $X'$ replaces all of this with ``a long drama film'', ``pretty quiet'', ``an ordinary congee place'', ``a plate of stir-fried winter melon'', ``the taste was nothing special'', removing the pragmatic mark of ``over-specific'' (lexis and ethos in $Y$ collapse together).
\item \textbf{The narrator's cognitive turn is replaced by a rationalising inference.} In the original the narrator goes through a stepwise cognitive collapse: ``This was really strange'' $\to$ ``Could it be that Shen is a ghost'' $\to$ ``but are there ghosts in this world?'' $\to$ ``maybe there has never been any Shen at all''. This is exactly the micro-instance of the chain rule in \S4: a macro decision pours strong constraint into thousands of words of $Y$, with $X_1$ suddenly crystallising the framework-resetting insight that ``Shen does not exist''. $X'$ replaces this with procedural rationalisations such as ``probably Shen had gone to the bathroom or was in line buying water'' and ``splitting up for a bit in the middle is a very normal thing'', pushing the cognitive turn back to the everyday default distribution; the dianoia-level reorganisation does not happen.
\item \textbf{The precise metaphor at the end is replaced by a state description.} The original ``her eyes shining, her face like a peach blossom, like a young female lead on an old opera stage''---the image of the ``young female lead on a stage'' precisely anchors ``performing'', and closes the semantic loop with ``daydream'' in the next paragraph. $X'$ replaces it with ``her face calm, her tone flat, as if she were just chatting about some very ordinary little daily thing'', cutting ``performance'' down to ``everyday'' and fully detaching from the ``daydream'' that just crystallised in the paragraph above.
\end{itemize}

After the degradation, $X'$ is still a Chinese narrative passage that is grammatically smooth and internally consistent, but its coupling with $Y$ drops sharply on all three dimensions of ethos, lexis and dianoia. $H(X' \mid Y)$ is about $0.86$ bit/token higher than $H(X \mid Y)$, and $\Delta I = +0.513$ bit/token---one of the largest gaps among the 20 pairs (see Table~\ref{tab:results}, row 12).

\end{document}